\documentclass[graybox]{svmult}

\usepackage{mathptmx}       
\usepackage{helvet}         
\usepackage{courier}        
\usepackage{type1cm}        
\usepackage{makeidx}         
\usepackage{graphicx}        
\usepackage{multicol}        
\usepackage[bottom]{footmisc}

\usepackage{amsmath}
\usepackage{float}
\usepackage{hyperref}
\usepackage{natbib}
\usepackage{subcaption}

\index{phylogenetic diversity, phenotypic diversity, diversity, useful diversity, lexicase selection, Eco-EA, fitness sharing, exploration diagnostic, causality, metrics, feedback loop}

\makeindex             

\begin{document}

\title*{What can phylogenetic metrics tell us about useful diversity in evolutionary algorithms?}
\titlerunning{Phylogeny metrics}

\author{Jose Guadalupe Hernandez, Alexander Lalejini, and Emily Dolson}
\institute{Jose Guadalupe Hernandez \at BEACON Center for the Study of Evolution in Action and Department of Computer Science and Ecology, Evolutionary Biology, and Behavior Program, Michigan State University, East Lansing, MI, USA \email{herna383@msu.edu} 
\and Alexander Lalejini \at Department of Ecology and Evolutionary Biology, University of Michigan, Ann Arbor, MI, USA \email{ amlalejini@gmail.com} 
\and Emily Dolson \at BEACON Center for the Study of Evolution in Action and Department of Computer Science and Ecology, Evolutionary Biology, and Behavior Program, Michigan State University, East Lansing, MI, USA \email{dolsonem@msu.edu} }
%
%
\maketitle


\abstract{It is generally accepted that ``diversity'' is associated with success in evolutionary algorithms. However, diversity is a broad concept that can be measured and defined in a multitude of ways. To date, most evolutionary computation research has measured diversity using the richness and/or evenness of a particular genotypic or phenotypic property. While these metrics are informative, we hypothesize that other diversity metrics are more strongly predictive of success. Phylogenetic diversity metrics are a class of metrics popularly used in biology, which take into account the evolutionary history of a population. Here, we investigate the extent to which 1) these metrics provide different information than those traditionally used in evolutionary computation, and 2) these metrics better predict the long-term success of a run of evolutionary computation. We find that, in most cases, phylogenetic metrics behave meaningfully differently from other diversity metrics. Moreover, our results suggest that phylogenetic diversity is indeed a better predictor of success.}

\section{Introduction}
\label{sec:introduction}

Maintaining a sufficiently diverse population to successfully solve challenging problems is a central challenge in all branches of evolutionary computation. If the population's diversity collapses, an evolutionary algorithm can prematurely converge on a sub-optimal solution from which it is unable to escape \citep{goldberg_genetic_1987}. 
While many diversity maintenance techniques have been designed to combat this challenge, we currently lack a clear understanding of what factors contribute to their success or failure in any given situation.
Broadly speaking, diversity maintenance techniques can fail in two ways:
1) failure to maintain a diverse population at all, and 2) failure to maintain diversity that is actually helpful to solving the problem. 
Because more effort has historically been paid to the former category, here we will focus on the latter.

\begin{figure}
\includegraphics[width=\linewidth]{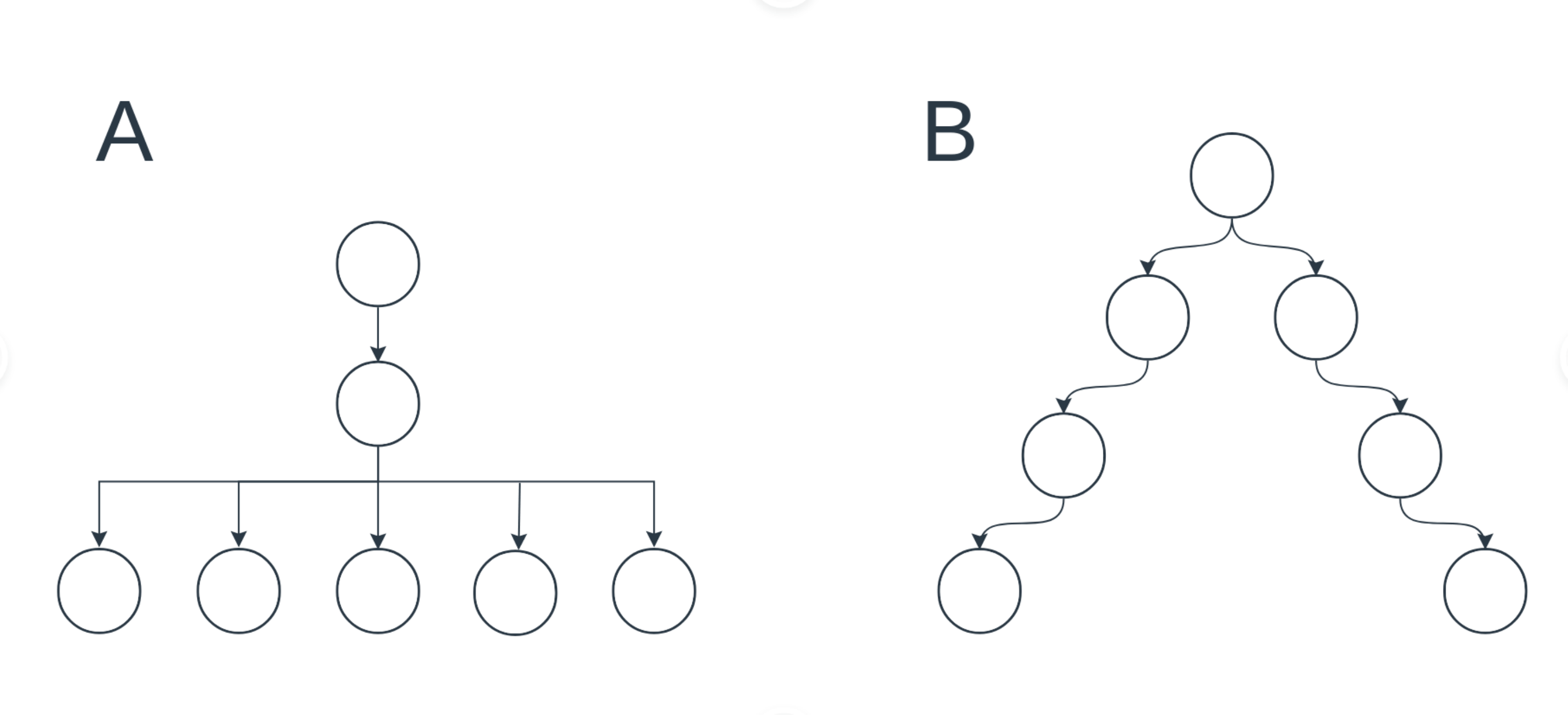}
\caption{\textbf{Example populations with different kinds of diversity}. This figure shows two different phylogenies. Arrows show parent-child relationships. Each node is a taxonomically unique phenotype (i.e., a phenotype with a unique evolutionary origin). For simplicity, leaf nodes in these diagrams are assumed to be the current set of taxa in the population; in reality, there could be non-leaf nodes corresponding to extant taxa. A) A population with high phenotypic diversity (phenotypic richness = 5) and low phylogenetic diversity (mean pairwise distance = 2). B) A population with low phenotypic diversity (phenotypic richness = 2) and high phylogenetic diversity (mean pairwise distance = 6).  }
\label{fig:conceptual}
\end{figure}

In evolutionary computation, diversity is usually evaluated by counting the number of unique ``types'' in the population.
These ``types'' may be genotypes, phenotypes, ecotypes, species, or other descriptors of a group of solutions.
The most commonly used types are phenotypes, which are often described using error vectors, behaviors, or output values.
Sometimes a population's diversity is measured as the raw count of unique types (in biology, such metrics are called ``richness''). Alternatively, sometimes different metrics are used that take into account how evenly the population is distributed across the types (as in Shannon diversity/entropy). Occasionally more nuanced metrics are used that consider the level of similarity of the types in the population (e.g., calculating Euclidean distance between error vectors, or including some sort of clustering step). However, the metrics commonly used in evolutionary computation are a small subset of the broader range of ways one could measure diversity. Prior work suggests that some diversity metrics are more predictive of success than others \citep{mouret_overcoming_2009, jackson_promoting_2010}, so investigating the implications of a wider variety of diversity metrics is worthwhile.

A variety of diversity metrics from ecology and evolutionary biology have yet to be examined in the context of evolutionary computation. One particularly interesting class of diversity metrics, called phylodiversity metrics \citep{tucker_guide_2017}, takes into account the evolutionary history of a population to calculate its diversity. For intuition behind how these metrics are different from other diversity metrics, see Figure \ref{fig:conceptual}. 
They do so by measuring the topology of the phylogenetic tree for the population (i.e., the ancestry tree). Intuitively,  types that are more evolutionarily distant from each other (i.e., share a more distant common ancestor) are likely farther apart from each other in the fitness landscape. 
As such, 
the ability to maintain evolutionarily distinct types may be important for effective problem-solving in evolutionary computing, due to the likely difficulty of re-evolving such distinct types from each other. 
Notably, biological simulations suggest that phylogenetic diversity provides different information about a population than a specific type of phenotypic richness (functional diversity) does; moreover, the extent of this difference varies across different scenarios \citep{tucker_relationship_2018}.

Preliminary data supports the hypotheses that 1) phylogenetic diversity captures information that other diversity metrics do not \citep{dolson_ecological_2018, tucker_relationship_2018}, and 2) phylogenetic diversity may be a better predictor of success in genetic programming than other diversity metrics \citep{dolson_ecological_2018}. Here, we investigate these hypotheses in the context of a range of problems and diversity maintenance techniques. Our findings provide further support for these two hypotheses. In particular, using a technique called causality analysis, we show that phylogenetic diversity is a stronger predictor of future fitness than phenotypic diversity is.



\section{Methods}
\label{sec:methods}

To identify phylogenetic metrics that correlate with success in evolutionary computation, we ran an evolutionary algorithm on a variety of problems using a range of parent selection methods in a full-factorial design. 
Specifically, we applied each of tournament, random, fitness sharing, lexicase, and Eco-EA selection to the following problems: an exploration diagnostic, Count Odds, Logic-9, sorting networks, and NK Landscapes.

\subsection{Selection methods}
\label{sec:methods:selection-methods}

\subsubsection{Tournament}

In tournament selection, $T$ individuals are randomly selected from the population to form a ``tournament''. The individual in the tournament that has the highest fitness is selected to reproduce. Tournament selection is included here as a control, as it is simple and introduces no pressure for diversity.

In the exploration diagnostic experiments, we used a tournament size of $T=8$ for consistency with \citep{hernandez_exploration_2021}. In the other experiments, we used a tournament size of $T=2$.

\subsubsection{Random}

Random selection is also included in this paper as a control. In this selection method, individuals are selected at random to reproduce.

\subsubsection{Fitness sharing}

Fitness sharing is one of the earliest diversity maintenance techniques \citep{goldberg_genetic_1987}. It is very similar to tournament selection, except that before the tournament occurs, the fitness of each individual is discounted in proportion to the number of similar individuals and the degree of their similarity. As a result, fitness sharing creates pressure for increased population diversity. 

Fitness sharing requires two parameters: $\alpha$ (which controls the shape of the sharing function) and $\sigma_sh$ (which controls the niche width). Here, we used $\alpha = 1$ and $\sigma_sh=2$.

\subsubsection{Lexicase selection}

Lexicase selection is designed to work in situations where fitness can be broken down in to multiple constituent parts \citep{spector_assessment_2012}. Conventionally, these constituent parts, or fitness criteria, are individual test cases in a genetic programming problem, but lexicase selection has been shown to work well in other scenarios too \citep{dolson_applying_2018, metevier_LexicaseSelectionGenetic_2019}. To select a parent with lexicase selection, the fitness criteria are randomly ordered. Then the criteria are stepped through in order, with all but the best performers on each criterion being eliminated from consideration for this selection event. This process continues either until only one individual remains to be selected or until there are no criteria left, in which case an individual is randomly selected from those remaining. Lexicase selection is known to generally maintain high levels of diversity \citep{helmuth_solving_2015, helmuth_lexicase_2016}, and preliminary evidence suggests that lexicase selection is also particularly effective at maintaining phylogenetic diversity \citep{dolson_ecological_2018}.

\subsubsection{Eco-EA}

Like lexicase selection, Eco-EA is designed to work on problems that have various sub-components \citep{goings_ecological_2009, goings_ecology-based_2012}. The original inspiration for Eco-EA came out of problems in which solutions to multiple simpler tasks served as helpful building blocks for solving a more complex overall task. These building block tasks are analogous to fitness criteria in lexicase selection \citep{dolson_applying_2018, dolson_ecological_2018}. Eco-EA associates each sub-task with a limited resource. Individuals that perform a task gain fitness while resources associated with that task are available; however, this decreases that task's associated resources. When the resource is depleted, there is no fitness benefit to performing the task. New resources flow in and out at a constant rate over the course of each update. As such, there is pressure for diversity.

The specific fitness gain associated with performing a task is described by the following function:

\begin{equation}
    \label{ecoeafitness}
    \text{total\_fitness} = \text{base\_fitness} * 2^{\text{min(score}^{\alpha}\text{ }*\text{ }C_f\text{ }*\text{ }\text{resource, max\_bonus})}
\end{equation}

where base fitness is the individual's fitness according to the global fitness function, score is the individual's score on the current sub-task, $\alpha$ is a variable that tunes the shape of the relationship between sub-task performance and fitness gain, $C_f$ is the fraction of available resources that can be consumed, resource is the amount of the resource available, and max bonus is the maximum allowed bonus. To strengthen the negative frequency dependence, a cost can be applied for attempting to use more resource than is available.

For the exploration diagnostic, we used $C_f = 1$, $\alpha=.25$, cost $=,1$, and max bonus $= 5$. Resources flow in at a rate of 250 units per generation. Every generation, .01\% of the total quantity of each resource flows out. For the other experiments, we used $C_f = .01$, $\alpha = 2$, cost $= 3$, and max bonus $= 5$. Resources flow in at a rate of 50 units per generation and 5\% of each resource flows out each generation.

\subsection{Problems}

\subsubsection{Exploration diagnostic}

Recently, Hernandez et al. proposed a suite of ``diagnostic'' fitness landscapes designed to test evolutionary algorithms in a controlled environment \citep{hernandez_exploration_2021}. Among these is an exploration diagnostic, which tests an algorithm's ability to simultaneously explore many different paths through a fitness landscape. In this diagnostic fitness landscape, genomes are vectors of floating point numbers. An individual's fitness is determined by finding the highest value in its genome, which represents the start of the values that contribute to fitness. This value and all subsequent monotonically decreasing values are summed together to produce the total fitness. For lexicase selection and Eco-EA, the fitness contribution of each individual site was used as the selection criterion or sub-task respectively.

Our hypotheses are predicated on the idea that phylogenetic diversity is a better measure of fitness landscape exploration than more conventionally used diversity metrics.
As such, the exploration diagnostic represents an ideal context in which to test our hypotheses. Moreover, this fitness landscape provides a smooth and controlled environment with more intuitive diversity dynamics. In contrast, more complex fitness landscapes can yield idiosyncratic crashes in diversity that vary based on the path(s) the population takes. Thus, we will use the exploration diagnostic for our in-depth analysis, and provide results in the more complex landscapes as a follow-up investigation of the robustness of our findings.

\subsubsection{Count Odds}

Count Odds is part of a benchmark suite of program-synthesis problems derived from a set of real programming problems given to introductory computer science students \citep{helmuth_general_2015}. These problems are specified using sets of testcases where each testcase contains inputs and expected outputs. Solutions to these problems consist of evolved code that produces the correct output for a given set of inputs. In the Count Odds problem, programs are given a list of numbers as input and must output an integer indicating how many of those numbers are odd.

For lexicase selection and Eco-EA, each test case is used as one selection criterion or sub-task, respectively. For fitness sharing, similarity is calculated based on the euclidean distance of the vector of performances on all test cases.

\subsubsection{Logic-9}

In Logic-9, a solution comprises code that, when executed, performs the nine one- and two-input bitwise Boolean logic tasks (\textit{not}, \textit{and}, \textit{or}, \textit{nand}, \textit{orn}, \textit{andn}, \textit{nor}, \textit{xor}, and \textit{equ}) \citep{ofria_avida:_2004, dolson_exploring_2019}.
To evaluate a program, we provide two numeric inputs that the program may use during its execution. Over the course of evaluation, a program may output numbers. We check whether each output is the result of performing one of the nine logic operations. If so, the program is counted as having completed that operation. Every logic operation completed increases the program's score by one. In addition to the Boolean logic tasks, programs get credit for solving the \textit{echo} task, in which the program must output one of the numbers it received as input. This task is included because it is known to be helpful in scaffolding the evolution of the more complex logic tasks.

For lexicase selection and Eco-EA, each logic task is used as one selection criterion or sub-task, respectively. For fitness sharing, similarity is calculated based on the euclidean distance of the vector of which tasks were completed.

\subsubsection{NK Landscapes}

In NK Landscapes, individuals are bitstrings of length $N$ \citep{kauffman_towards_1987}. 
Each position in the bitstring has an associated table used to look up the fitness contribution of that position, which is based on the value (0 or 1) at that position and the values of $K$ neighboring positions.
Thus, each lookup table comprises $2^{K+1}$ entries, one for each possible sequence of relevant bits.
Total solution quality for a given bitstring individual is calculated by summing up the fitness contributions from each site. 

An NK Landscape is generated by randomly selecting a value between 0 and 1 to be the fitness contribution for each entry in each site's fitness contribution lookup table.
$N$ determines the size of the fitness landscape, and $K$ determines the landscape's level of epistasis or ruggedness (i.e., how interdependent the fitness contribution of each site is on other sites).
For example, if $K$ is 0, each site has two possible fitness contributions, one for if that position is set to 1 and one for if it is set to 0. If $K$ is set to 1, each site has four possible fitness contributions based on the value of that site and the value of its neighbor (one each for 00, 01, 10, and 11). Every increase in $K$ doubles the possible different fitness contributions. 

For lexicase selection and Eco-EA, the fitness contribution from each site is used as one selection criterion or sub-task, respectively. For fitness sharing, similarity is based on the Hamming distance between the solutions. For the experiments presented here, we used $K=3$ and $N=20$.

\subsubsection{Sorting Networks}

Sorting networks are computational units designed to sort fixed-length sequences of numbers via a set of comparisons between pre-specified positions in the sequence \citep{sekanina_EvolutionaryDesignArbitrarily_2005}. When the network compares two positions, the numbers in them are swapped if they are out of order. In this problem, individuals are represented as sequences of comparators between positions. As with the Count Odds problem, fitness is assessed via a sequence of test cases. Once all test cases are solved correctly, individuals can receive additional fitness bonuses for having as few comparators as possible. For lexicase selection and Eco-EA, each test case is used as a single selection criterion or sub-task respectively.

Here, we evolved sorting networks to sort 30 values and test them on 100 test cases. The maximum allowed number of comparators per network was 128.

\subsection{Computational Substrates}

For the genetic programming problems (Count Odds and Logic-9), we evolved linear genetic programs where each genome is a sequence of simple computational instructions. Most notably, the instruction set is designed to support the evolution of modularity by supporting the encapsulation of subroutines into ``scopes''. Programs have a set of read-only memory spaces used to provide input and a set of write-only memory spaces to be used as output. Each instruction in the genome is executed in sequence. If execution reaches the end of the genome before the program runs out of evaluation time, execution will loop back around to the front of the genome. We propagated programs asexually and applied the following mutations to offspring, each with a probability of 0.005: 1) instruction substitutions, 2) point insertions and deletions, and 3) substituting the argument being supplied to an instruction. This linear genetic programming representation is described in more detail in \citep{dolson_exploring_2019}.

For NK Landscapes, we evolved bitstrings. We propagated individuals asexually, and we applied bit flip mutations at a per-bit rate of 0.01.

In the exploration diagnostic, we evolved vectors of 50 floating point values where each value ranged between 0.0 and 25.0. 
We reduced the size of these ranges from those used by \cite{hernandez_exploration_2021} to ensure that Eco-EA had sufficient time to solve the problem within our allocated computational budget. 
We used asexual reproduction, and each position had a 0.007 probability of mutating. Mutations modify a value by a number drawn from a normal distribution with mean 0 and standard deviation 1.

Sorting networks are represented as sequences of pairwise comparators. They can mutate via insertions of new comparators (0.0005 probability), duplications of existing comparators (0.0005 probability), deletions of existing comparators (0.001 probability), swapping pairs of existing comparators (0.001 probability), and substituting different indices in existing comparators (0.001 probability).

\subsection{Other Parameters}

For the exploration diagnostic, we used a population size of 500 and allowed runs to evolve for 500,000 generations. This length was selected to ensure that all selection schemes (most notably Eco-EA) had adequate time to find a good solution. For the other fitness landscapes, we used a population size of 1000 and allowed runs to evolve to 1,000 generations.

\subsection{Phylogenetic Diversity Metrics}
\label{section:metrics}

A wide variety of phylogenetic diversity metrics have been developed, and the extent to which they capture different information from each other is an area of active research \citep{tucker_guide_2017}. All of them require that you keep track of the full phylogeny (ancestry tree) of a population. For further discussion of building phylogenies in the context of evolutionary computation, see \citep{dolson_interpreting_2020}. Here, we focus on two classes of metrics: pairwise distance metrics and evolutionary distinctiveness metrics. 

Pairwise distance metrics calculate the number of edges\footnote{Weighted edges can also be used, in which case the weights along the path should be summed. Here, we use unweighted edges.} in the shortest path between each pair of nodes associated with extant taxa (i.e., taxonomic units corresponding to individuals in the current population) \citep{webb_phylogenies_2002}. The resulting set of distances can then be summarized by calculating its minimum, maximum, mean, or variance. Each of these statistics produces a different phylogenetic diversity metric with different properties.

Evolutionary distinctiveness metrics assign an evolutionary distinctiveness score to each extant taxon \citep{isaac_mammals_2007}. This score takes into account each branch's age, and represents how evolutionarily distant each taxon is from all other extant taxa. To calculate evolutionary distinctiveness, the age of each branch is calculated and divided by the number of extant taxa the branch ultimately leads to. A taxon's evolutionary distinctiveness is the sum of the values calculated for all branches between that taxon and the tree's root. As with pairwise distances, the set of evolutionary distinctiveness scores can be summarized by taking its minimum, maximum, mean, or variance.

\subsection{Analysis techniques}

\subsubsection{Statistics}

Correlations among variables at a fixed time point were measured using Spearman correlations. We used Spearman correlations rather than Pearson correlations due to the fact that many of the relationships being measured were non-linear (but still monotonically increasing). To compare different conditions, we used Kruskal-Wallis tests with subsequent pairwise Wilcoxon rank-sum tests and a Bonferonni correction for multiple comparisons. To ensure statistical rigor, we decided what set of statistical comparisons to perform based on analysis of an initial pilot data-set. We then re-ran all experiments with different random seeds and performed the pre-determined analysis on this new data to generate the results presented here. 

\subsubsection{Transfer entropy}

As alluded to previously, there is a positive feedback loop between diversity and evolutionary success during the initial phase of evolution. To further complicate matters, fully solving a problem can lead diversity to crash as that solution sweeps through the population. For these reasons, looking at correlations between fitness and diversity at any given time point gives us only limited information about their relationship. One approach to getting around this problem is to take all measurements at the time step when a perfect solution is first discovered \citep{dolson_SpatialResourceHeterogeneity_2017a}. While this approach helps, it does not address the question of what is driving the feedback loop. Moreover, even before a perfect solution evolves, the evolution of partial solutions may initiate partial selective sweeps.

In order to understand the specific role that different types of diversity play in driving the feedback loop between diversity and success, we turn to an analytical approach called causality analysis. As the nature of causality can quickly drift into murky philosophical territory, here we specifically use a notion of causality called Granger causality \citep{granger_InvestigatingCausalRelations_1969, bressler_WienerGrangerCausality_2011}. We say that $X$ Granger-causes $Y$ if past values of $X$ contain information about the current value of $Y$ above and beyond the information that past values of $Y$ contain about the current value of $Y$. This definition comes from the insights that 1) the past causes the future, not the other way around, and 2) if $X$ and $Y$ are jointly caused by an external process, that process will be captured by the information that past values of $Y$ have about the current value of $Y$.

Granger causality is normally measured in the context of vector auto-regressive of models. However, our data do not match the assumptions of such models (particularly stationarity). Thus, we measure Granger-causality with an information theoretic metric called Transfer Entropy \citep{schreiber_MeasuringInformationTransfer_2000, yao_EffectiveTransferEntropy_2020}. In information theoretic terms, transfer entropy is $I(Y_{t} ; X_{t-k}|Y_{t-k})$, the conditional mutual information between $Y_{t}$ and $X_{t-k} | y_{t-k}$. Here, $Y$ is the variable being predicted, $t$ is the time point it is being predicted at, $X$ is the variable we are using the predict $Y$, and $k$ is the ``lag''. The lag specifies the time scale on which we are interested in measuring Granger causality. In other words, it indicates which past value of $X$ we are attempting to use to predict the current value of $Y$. 

Often, the goal of these measurements is simply to establish the direction of Granger-causality. In those cases, very short lags are often used. In this case, however, it would also be valuable to know whether predictive capability is maintained over large lags. If it were, there would be a variety of useful practical implications. For example, we could potentially use the diversity at a relatively early time point to predict whether a given run of evolutionary computation will be successful. For this reason, we measure Transfer Entropy using lags ranging from 10 generations to 100,000 generations. 

Here, we measure both the Transfer Entropy between fitness and phylogenetic diversity and the Transfer Entropy between fitness and phenotypic diversity. Note that, because Transfer Entropy is a value calculated based on two time series, we cannot meaningfully lump multiple replicates into the same calculation. Thus, for each condition we will end up with a distribution of Transfer Entropy values.

\subsection{Code Availability}

All code used in this paper is open source and freely available in the supplemental material \citep{supplement}. Research code was written in C++ using the Empirical library \citep{ofria_empirical_2018}. Data analysis was performed using the R statistical computing language, version 4.0.4 \citep{r2021}, the ggplot2 \citep{wickham_ggplot2:_2016}, ggpubr \citep{ggpubr}, and infotheo \citep{infotheo} libraries.

A C++ implementation of phylogeny tracking and all phylodiversity metrics used here is available in the Empirical library \citep{ofria_empirical_2018}. This implementation is designed to plug into any computational evolution code written in C++.

\section{Results and Discussion}
\label{sec:results}

\subsection{Do phylogenetic metrics provide novel information?}

If phylogenetic diversity is to tell us anything useful, a necessary first step is that it provide information that more commonly-used metrics do not. It is particularly important to establish this distinction in light of the fact that phylogenetic metrics are typically more computationally expensive to calculate.

\begin{figure}
\includegraphics[width=\linewidth]{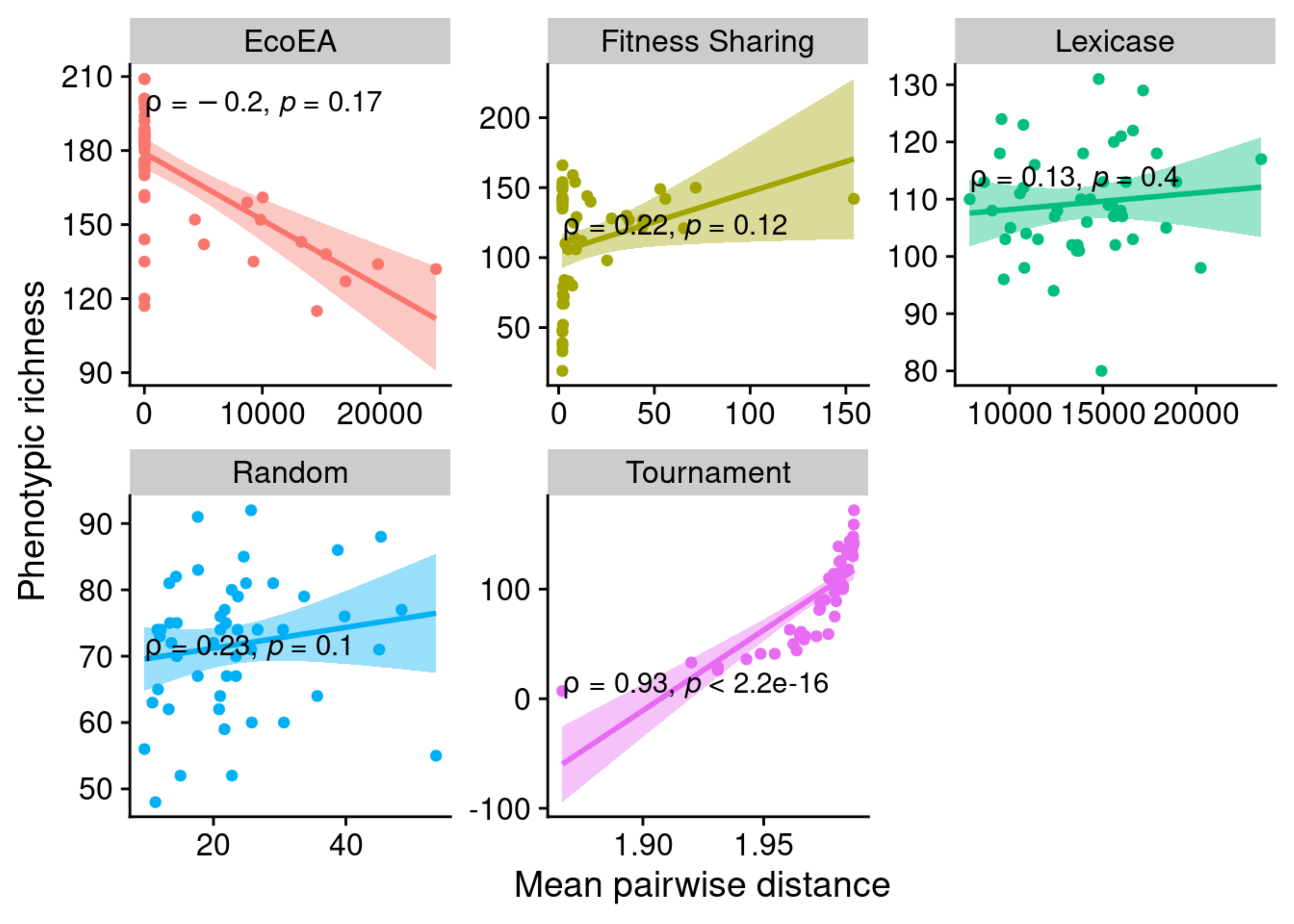}
\caption{\textbf{Relationship between phenotypic diversity and phylogenetic diversity at a single time point for the exploration diagnostic}. Values were measured at the final time point (generation 500,000). Spearman correlation coefficients are shown for each selection scheme, along with a 95\% confidence interval around regression lines. Note that axis scales vary between panels.}
\label{fig:phenotypic_vs_phylogenetic}
\end{figure}

First, we investigated the relationships between different metrics of phenotypic and phylogenetic diversity. The two measurements of phenotypic diversity that we analyzed were phenotypic richness and phenotypic Shannon diversity, the two most commonly used diversity metrics (see supplemental material \citep{supplement}). 
Unsurprisingly, phenotypic richness and phenotypic Shannon diversity are closely correlated across all conditions. We performed the rest of the analyses in this paper using both richness and Shannon diversity, but observed qualitatively the same results for both. For simplicity, we only present the results using richness (for Shannon diversity results, see supplemental material \citep{supplement}).

There are many ways of measuring phylogenetic diversity \citep{tucker_guide_2017}.
Here, we focus on the pairwise distance and evolutionary distinctiveness metrics (see \ref{section:metrics} for more information). Minimum pairwise distance was not informative, as there are nearly always at least two taxa a distance of 1 away from each other. In the conditions observed here, the other pairwise distance metrics tended to correlate fairly closely with each other. Correlations among evolutionary distinctiveness metrics were weaker and less consistent, but still present in most cases. In contrast, we did not observe consistent relationships between pairwise distance metrics and evolutionary distinctiveness metrics. While there were sometimes strong correlations within a condition, the direction of these correlations varied. For this reason, we performed subsequent analyses using both mean pairwise distance and mean evolutionary distinctiveness. As we observed qualitatively the same results for both, here we present only the results using mean pairwise distance (for evolutionary distinctiveness results, see supplemental material \citep{supplement}).


Next, we compared the phenotypic metrics to the phylodiversity metrics.  Intuitively, we might expect them to be highly correlated. In practice, however, we see that the correlation between phenotypic and phylogenetic diversity is not consistently\footnote{The correlation for tournament selection in the exploration diagnostic is incredibly high, however 1) the observed range of mean pairwise distance is so low that the correlation is almost certainly an artifact, and 2) this correlation is not observed for other fitness landscapes.} significantly different from 0 (see Figure \ref{fig:phenotypic_vs_phylogenetic}). In some cases, the correlation is even negative.



\begin{figure}
\centering
\begin{subfigure}[b]{\linewidth}

\includegraphics[width=.9\linewidth]{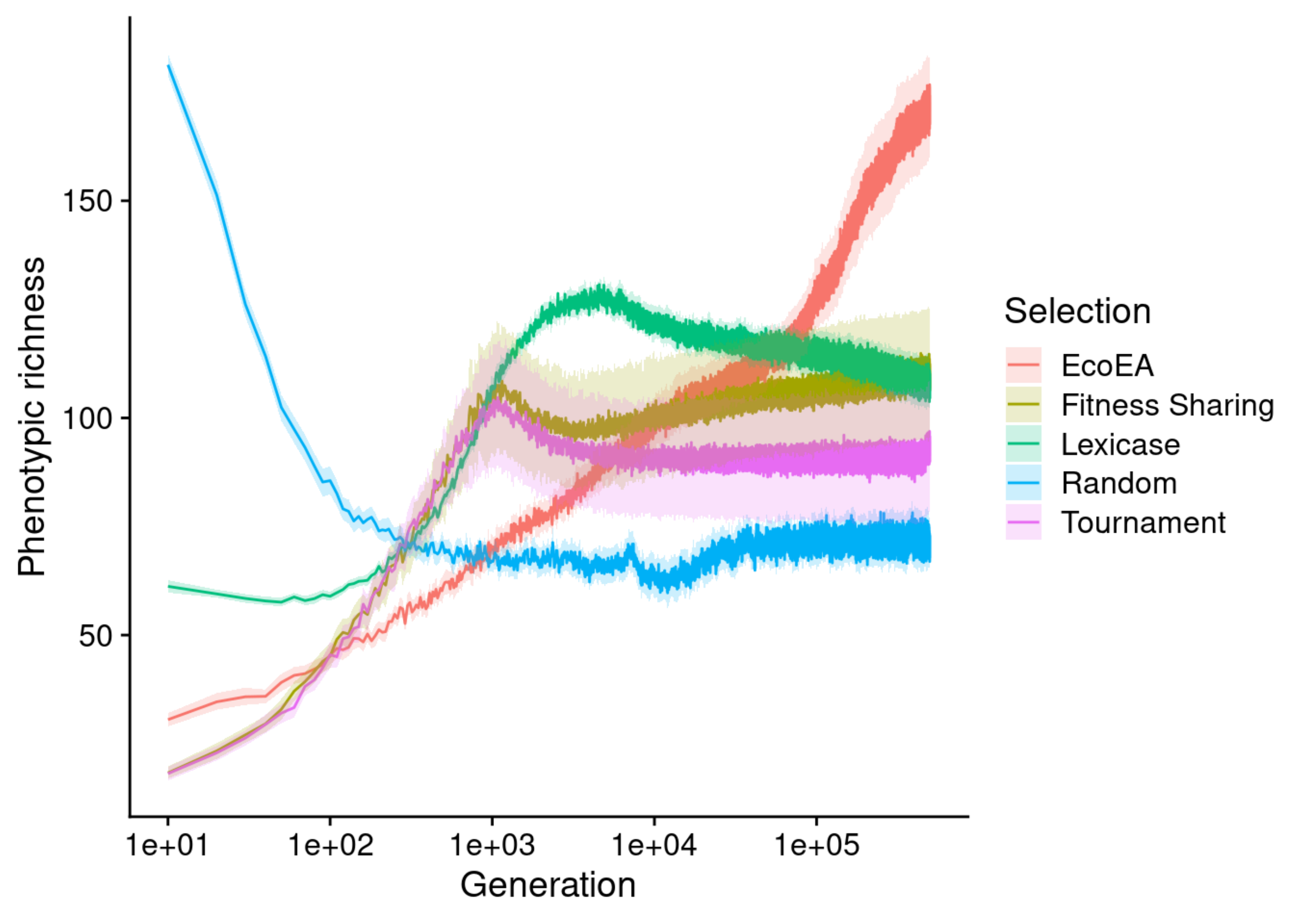}
\caption{\textbf{Phenotypic richness over time for each selection scheme on the exploration diagnostic}. Shaded areas represent 95\% confidence interval around the mean for all 50 replicates for each selection scheme. Note that the x-axis is on a log scale.}
\label{fig:phenotypic_over_time}
\end{subfigure}

\begin{subfigure}[b]{\linewidth}
\includegraphics[width=.9\linewidth]{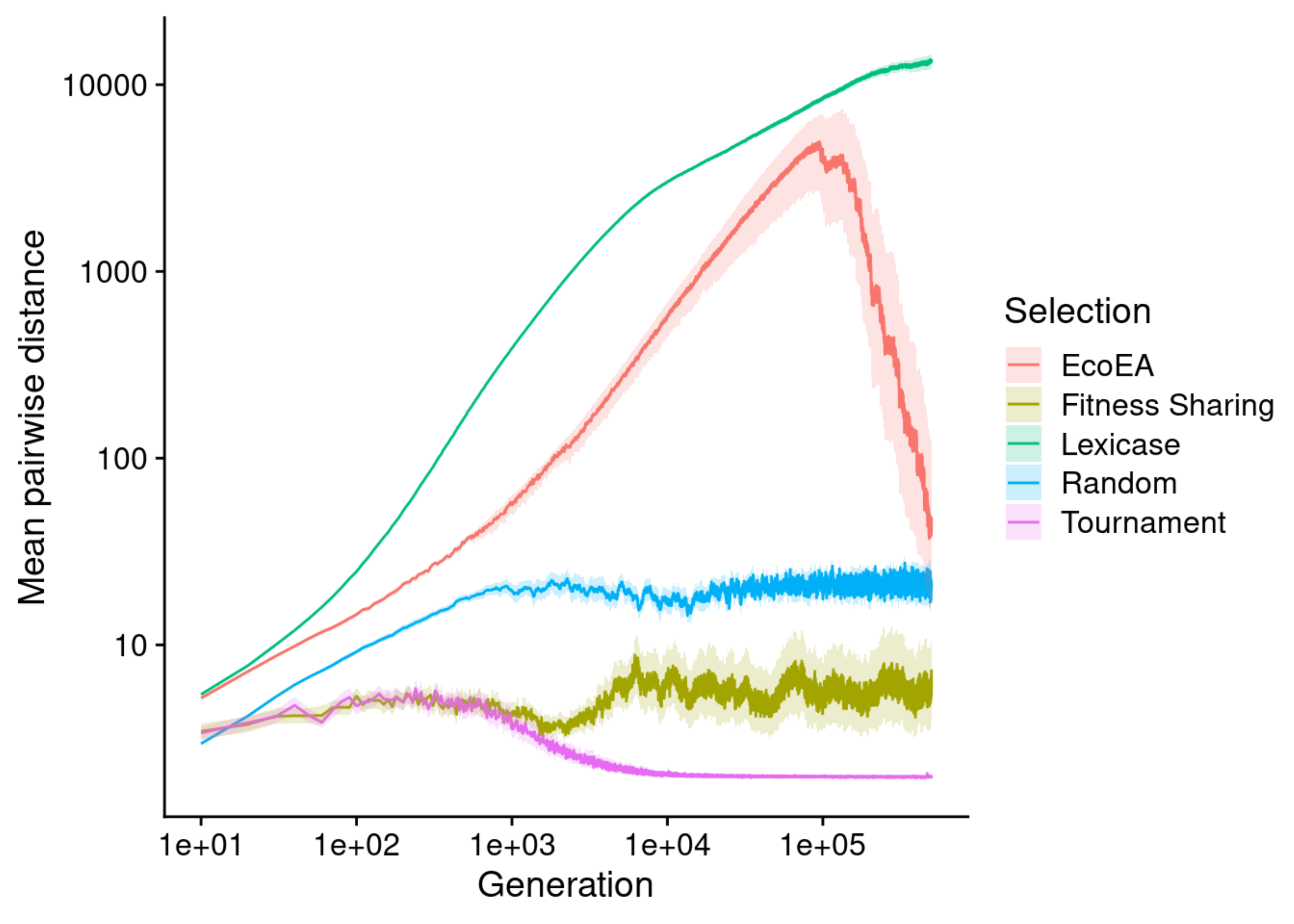}
\caption{\textbf{Phylogenetic diversity (mean pairwise distance) over time for each selection scheme on the exploration diagnostic}. Shaded areas represent 95\% confidence interval around the mean for all 50 replicates for each selection scheme. Note that the x-axis is on a log scale.}
\label{fig:phylogenetic_over_time}
\end{subfigure}
\end{figure}

As previously noted, diversity metrics can be sensitive to the exact time at which they are measured. To confirm that this lack of instantaneous correlation is indicative of a consistent lack of relationship, we plotted phenotypic and phylogenetic diversity over time (see Figures \ref{fig:phenotypic_over_time} and \ref{fig:phylogenetic_over_time}). Indeed, phenotypic and phylogenetic diversity behave differently over long temporal scales as well.

\begin{figure}
\includegraphics[width=\linewidth]{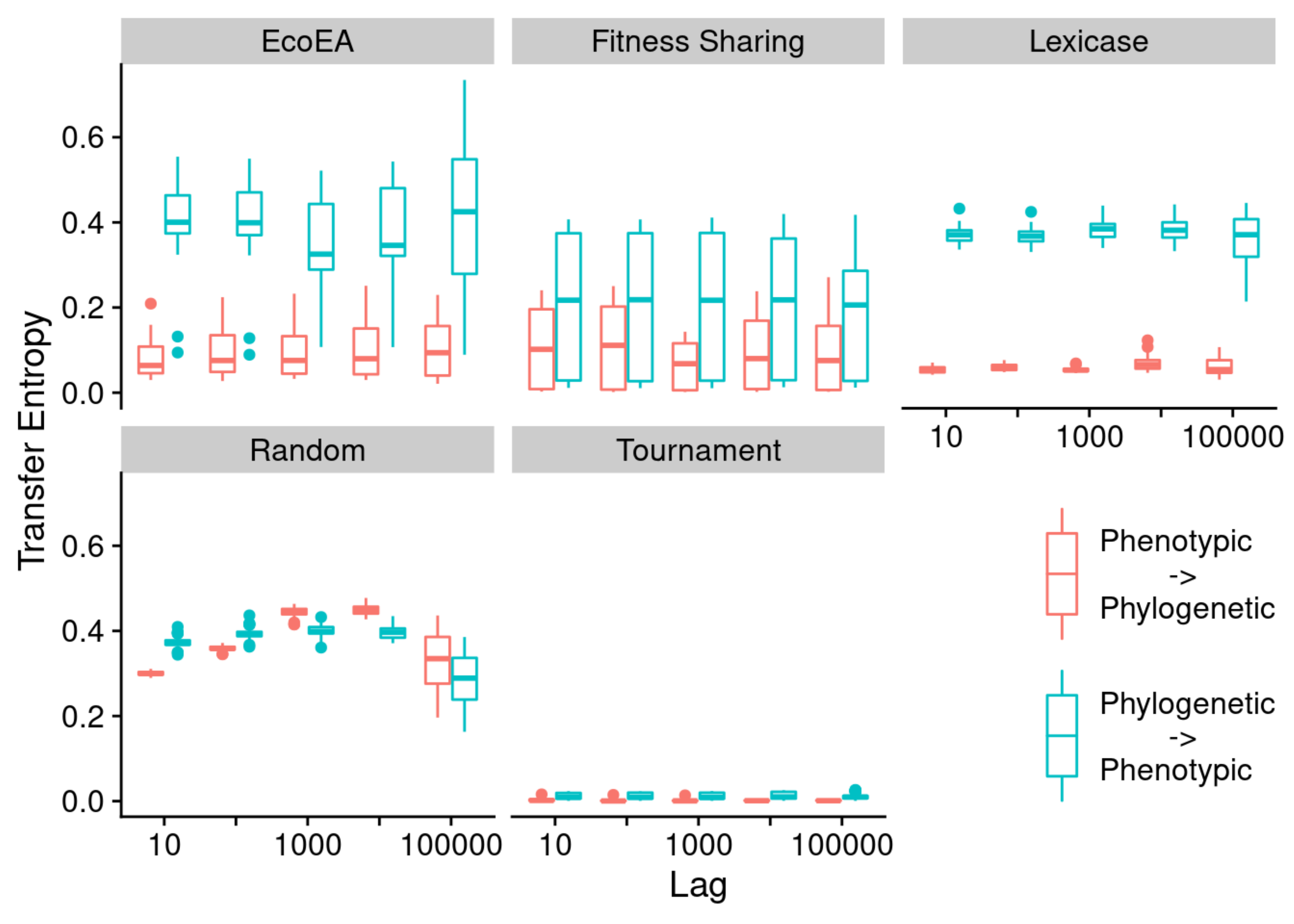}
\caption{\textbf{Transfer entropy between phenotypic and phylogenetic diversity on the exploration diagnostic}. Each boxplot shows the distribution of observed transfer entropies for each direction of transfer, lag, and selection scheme. Results shown here are for 50 replicate runs of each selection scheme on the exploration diagnostic. Note that the x-axis is on a log scale.}
\label{fig:te_diversity}
\end{figure}

Having established that phenotypic and phylogenetic diversity are not reliably correlated, 
we next asked whether either diversity metric can predict the other. Based on our hypothesis that phylogenetic diversity is useful because it more directly indicates a population's spread across the fitness landscape, we might expect current phylogenetic diversity to predict future phenotypic diversity. We tested this hypothesis by measuring the transfer entropy from phylogenetic diversity to phenotypic diversity. 
Consistent with our hypothesis,
transfer entropy from phylogenetic diversity to phenotypic diversity was generally higher than transfer entropy from phenotypic diversity to phylogenetic diversity (see Figure \ref{fig:te_diversity}). Thus, in the Granger sense of causality, we can say that phylogenetic diversity produces phenotypic diversity to a greater extent than phenotypic diversity produces phylogenetic diversity.

In the exploration diagnostic fitness landscape, the only exceptions to this observation are our two control selection schemes: tournament selection and random selection. In tournament selection, neither form of diversity is particularly predictive of the other. This behavior is unsurprising, as tournament selection generally maintains minimal levels of both forms of diversity. In random selection, both forms of diversity are somewhat predictive of each other. In the other fitness landscapes we observe more variation in transfer entropy (see supplemental material \citep{supplement}).

Taken together, these results provide strong evidence that, in an evolutionary computation context, phylogenetic diversity metrics provide information that phenotypic diversity metrics do not. We base this conclusion on the lack of consistent correlation between these metrics at a fixed point in time, the differences in their long-term behavior, and that fact that the transfer entropy from phylogenetic diversity to phenotypic diversity is higher than the other way around (implying that phylogenetic diversity contains information about future phenotypic diversity that current phenotypic diversity does not contain about future phylogenetic diversity). We may now proceed to ask whether that information is actually useful for understanding problem-solving success in evolutionary computation.

\subsection{Do phylogenetic metrics predict problem-solving success?}

\begin{figure}
\includegraphics[width=\linewidth]{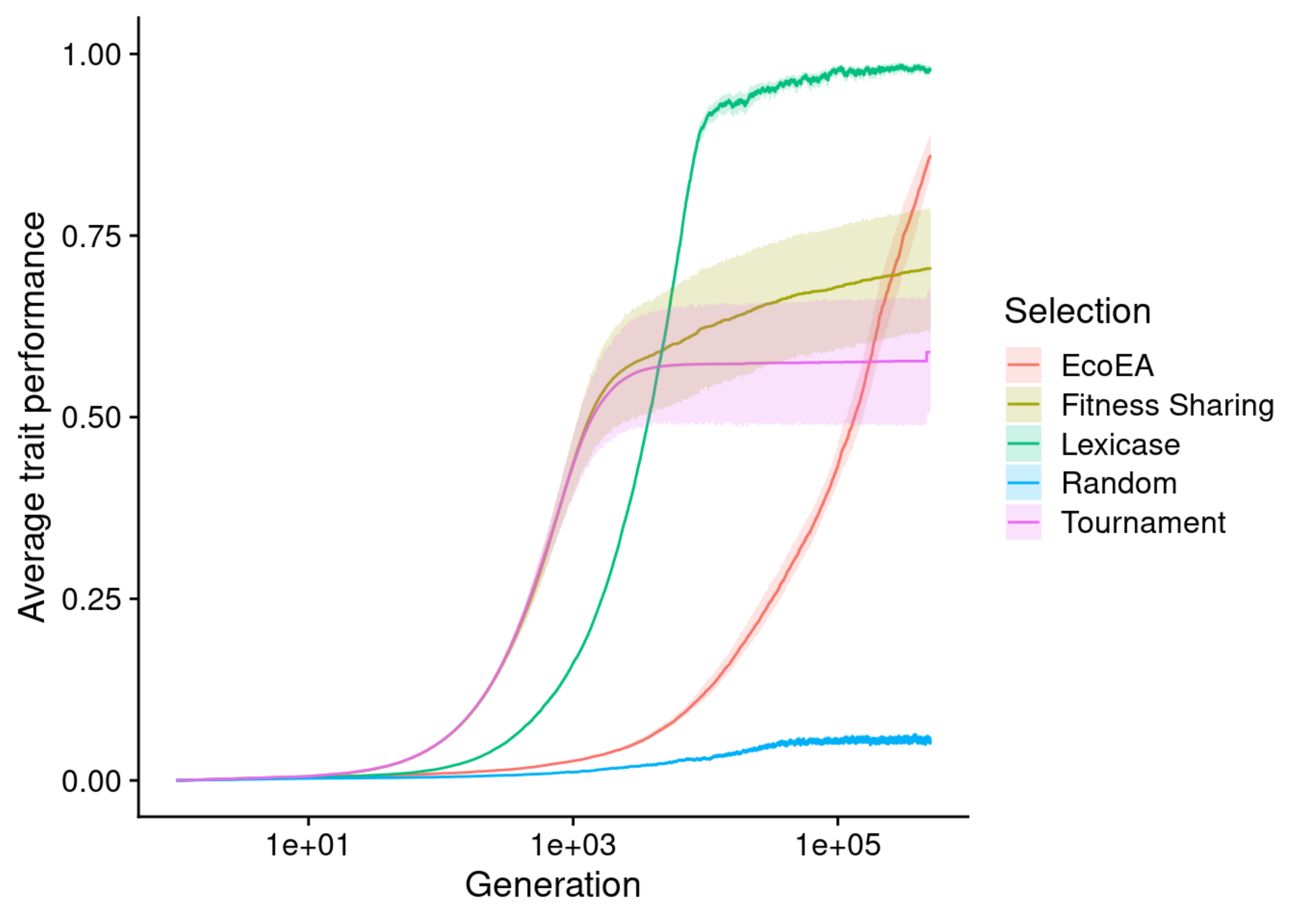}
\caption{\textbf{Fitness over time for each selection scheme on the exploration diagnostic}. Shaded areas represent 95\% confidence interval around the mean for all 50 replicates for each selection scheme. Note that the x-axis is on a log scale. The y-axis shows the proportion of the maximum possible fitness achieved.}
\label{fig:success_over_time}
\end{figure}

Having established that phylogenetic metrics provide different information than phenotypic diversity metrics, the next question to ask is what that information can tell us. In the exploration diagnostic landscape, we can see some intuitive connections between both types of diversity and fitness (see Figure \ref{fig:success_over_time}). Excluding random selection, the final performance of a selection scheme appears to be correlated with the final level of phylogenetic diversity maintained by that selection scheme (but not with the level of phenotypic diversity). In the other fitness landscapes, however, the connection between a selection scheme's ability to maintain diversity of either type and its ability to succeed is less obvious (see supplemental material \citep{supplement}). This increased complexity is unsurprising, as success in the exploration diagnostic landscape is based primarily on the ability to explore; succeeding on the other fitness landscapes is more complicated.

It should also be noted that, while there is a correlation between phylogenetic diversity and fitness in the aggregate on the exploration diagnostic landscape, there is not a consistent correlation among individual runs within a selection scheme (see supplemental material \citep{supplement}).
Eco-EA, for instance, appears to lose phylogenetic diversity as it approaches a good solution, leading to a negative correlation between fitness and phylogenetic diversity. For most of the other selection schemes, there is no significant correlation \footnote{In the pilot data set, we observed a strong positive correlation between phylogenetic diversity and fitness for lexicase selection. However, this correlation disappeared when we re-ran the experiments to generate the final data set.}.

\begin{figure}
\includegraphics[width=\linewidth]{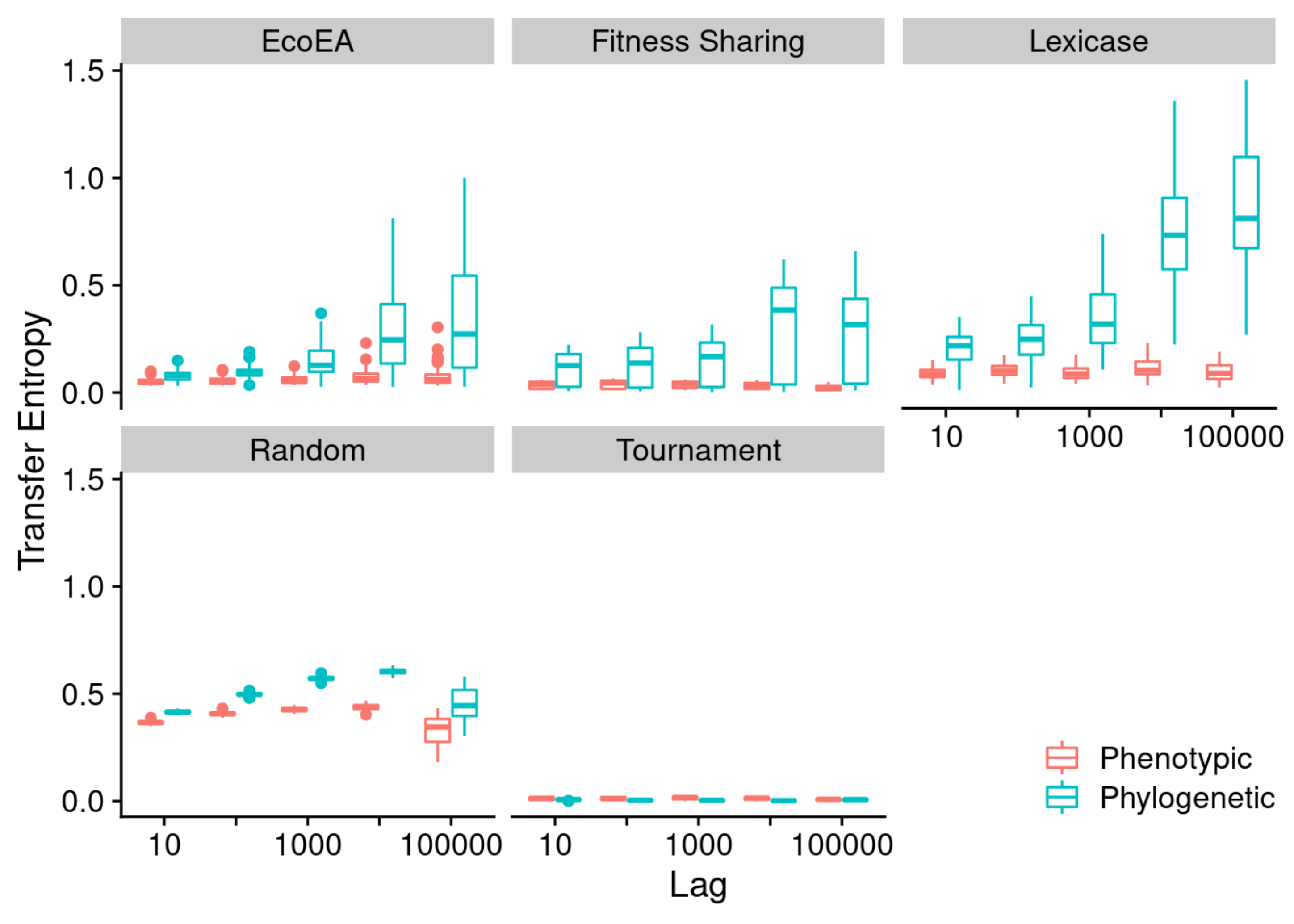}
\caption{\textbf{Transfer entropy from diversity to fitness for the exploration diagnostic}. Each boxplot shows the distribution of observed transfer entropies for each type of diversity, lag, and selection scheme. Results shown here are for 50 replicate runs of each selection scheme on the exploration diagnostic. Note that the x-axis is on a log scale.}
\label{fig:te_success}
\end{figure}

\begin{figure}
\includegraphics[width=\linewidth]{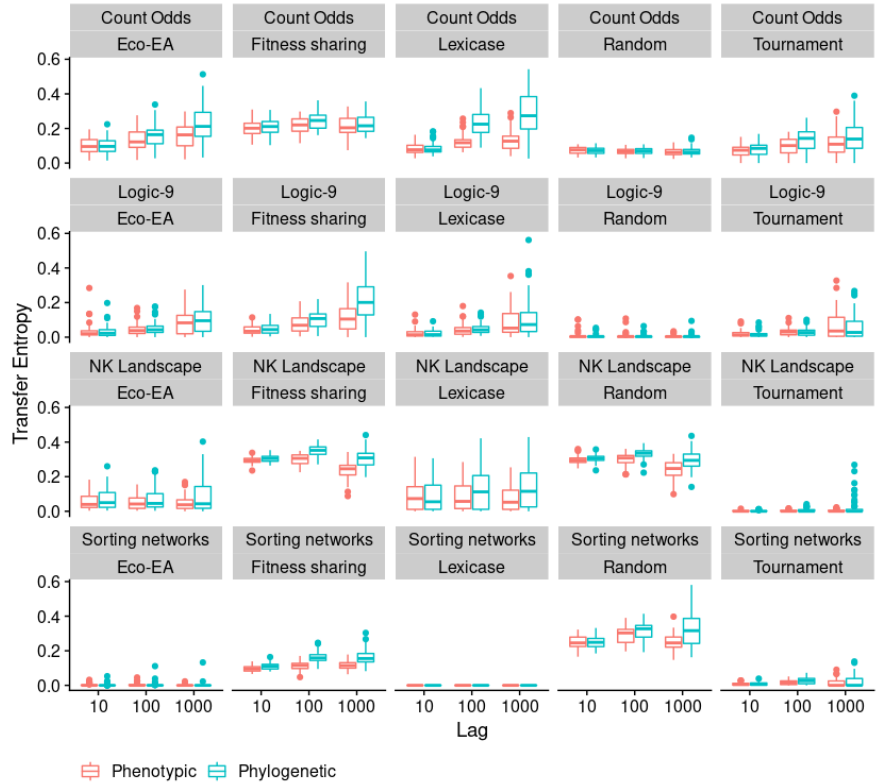}
\caption{\textbf{Transfer entropy from diversity to fitness for complex fitness landscapes}. Each boxplot shows the distribution of observed transfer entropies for each type of diversity, lag, and selection scheme. Results shown here are for 50 replicate runs of each selection scheme on each landscape. Note that the x-axis is on a log scale.}
\label{fig:te_success_all}
\end{figure}

To understand the precise dynamics driving the relationship between diversity and success, we measured the transfer entropy of phylogenetic diversity to fitness and the transfer entropy of phenotypic diversity to fitness. In the exploration diagnostic landscape, we see that, for the three non-control selection schemes, phylogenetic diversity is substantially more predictive of future fitness than phenotypic diversity (see Figure \ref{fig:te_success}). The predictive power of both types of diversity weakens substantially in the other fitness landscapes (see Figure \ref{fig:te_success_all}). However, when there is a discernible difference, phylogenetic diversity has higher transfer entropy than phenotypic diversity.

From these results, we can conclude that, for selection schemes that maintain diversity, phylogenetic diversity ``Granger-causes'' success to a greater extent than phenotypic diversity does. We interpret this as strong evidence that phylogenetic diversity is, in general, more predictive of success than phenotypic diversity (as measured by phenotypic richness and phenotypic Shannon diversity).





\section{Conclusion}
\label{sec:conclusions}

We have demonstrated that, in the context of evolutionary computation, phylogenetic diversity metrics capture information information that is substantially different from the information captured by conventionally used phenotypic diversity metrics (phenotypic richness and phenotypic Shannon diversity). The extent of this difference appears to vary by problem and by selection scheme, but it is evident in 1) the lack of consistent strong correlation between phenotypic and phylogenetic diversity at a fixed point in time, 2) the lack of similarity in long-term trends in phenotypic and phylogenetic diversity, and 3) the fact that phylogenetic diversity is better able to predict future phenotypic diversity than the other way around. 

Moreover, our results also suggest that phylogenetic diversity is generally a stronger driver of success than phenotypic diversity (in an evolutionary computing context). This finding also varies by selection scheme and by problem, with selection schemes that maintain diversity showing a stronger relative effect of phylogenetic diversity than tournament selection does. Impressively, this is true even at very long time lags; phylogenetic diversity provides predictive information about fitness tens of thousands of generations in the future.

Taken together, these results suggest that it may be worthwhile for researchers studying diversity in evolutionary computation to measure phylogenetic diversity in addition to or instead of phenotypic diversity. Doing so will take us a step closer to identifying diversity that is helpful to solving a given problem. Additionally, these results hint at the possibility of using phylogenetic diversity early in a run of evolutionary computation as a predictor of which runs will go on to be most successful. Anecdotally, we have found phylogenetic diversity in just the first few generations to be a useful indicator of whether we have correctly selected parameters for fitness sharing and Eco-EA.

In this paper, we have barely scratched the surface of phylogenetic metrics that may be relevant to evolutionary computation. Given our findings thus far, a more thorough investigation of other phylogeny metrics in the context of evolutionary computation is warranted. 
Additionally, future work should evaluate the relationships between other conventionally used approaches to measuring diversity (e.g., genetic diversity).
As we improve our understanding of these relationships, we may even be able to use them to make inferences about fitness landscapes \citep{dolson_interpreting_2020}. 
We hope that the results presented here will inspire others to incorporate a phylogenetic perspective into their evolutionary computation research. 

\section{Author contributions}

ED conceptualized the questions and experiments in this chapter, wrote the code for the non-exploration-diagnostic experiments, ran all experiments, analyzed the data, and wrote the first draft of this chapter. JGH and AL wrote the code for the exploration diagnostic experiments and assisted with the data analysis and writing. 

\section{Acknowledgements}
We thank members of the MSU ECODE lab, the MSU Digital Evolution lab, and the Cleveland Clinic Theory Division for the conversations that inspired this work. This research was supported by the National Science Foundation (NSF) through the BEACON Center (Cooperative Agreement DBI-0939454). Michigan State University provided computational resources through the Institute for Cyber-Enabled Research. Any opinions, findings, and conclusions or recommendations expressed in this material are those of the author(s) and do not necessarily reflect the views of the NSF, UM, or MSU.

%
%
\bibliographystyle{apalike}
\bibliography{references}

\begin{thebibliography}{}

\bibitem[Bressler and Seth, 2011]{bressler_WienerGrangerCausality_2011}
Bressler, S.~L. and Seth, A.~K. (2011).
\newblock Wiener–{Granger} {Causality}: {A} well established methodology.
\newblock {\em NeuroImage}, 58(2):323--329.

\bibitem[Dolson, 2021]{supplement}
Dolson, E. (2021).
\newblock {Supplemental material for "What can phylogenetic metrics tell us
  about useful diversity in evolutionary algorithms?" at GPTP 2021}.
\newblock {DOI: 10.5281/zenodo.4733407. URL:
  https://github.com/emilydolson/ecology\_in\_EC\_parameter\_sweep}.

\bibitem[Dolson et~al., 2018a]{dolson_applying_2018}
Dolson, E., Banzhaf, W., and Ofria, C. (2018a).
\newblock Applying {Ecological} {Principles} to {Genetic} {Programming}.
\newblock In Banzhaf, W., Olson, R.~S., Tozier, W., and Riolo, R., editors,
  {\em Genetic {Programming} {Theory} and {Practice} {XV}}, pages 73--88.
  Springer International Publishing, Cham.

\bibitem[Dolson et~al., 2020]{dolson_interpreting_2020}
Dolson, E., Lalejini, A., Jorgensen, S., and Ofria, C. (2020).
\newblock Interpreting the {Tape} of {Life}: {Ancestry}-based {Analyses}
  {Provide} {Insights} and {Intuition} about {Evolutionary} {Dynamics}.
\newblock {\em Artificial Life}, 26(1):1--22.

\bibitem[Dolson et~al., 2019]{dolson_exploring_2019}
Dolson, E., Lalejini, A., and Ofria, C. (2019).
\newblock Exploring {Genetic} {Programming} {Systems} with {MAP}-{Elites}.
\newblock In Banzhaf, W., Spector, L., and Sheneman, L., editors, {\em Genetic
  {Programming} {Theory} and {Practice} {XVI}}, pages 1--16. Springer
  International Publishing, Cham.

\bibitem[Dolson et~al., 2017]{dolson_SpatialResourceHeterogeneity_2017a}
Dolson, E., Perez, S., Olson, R., and Ofria, C. (2017).
\newblock Spatial resource heterogeneity increases diversity and evolutionary
  potential.
\newblock {\em bioRxiv}.

\bibitem[Dolson et~al., 2018b]{dolson_ecological_2018}
Dolson, E.~L., Banzhaf, W., and Ofria, C. (2018b).
\newblock Ecological theory provides insights about evolutionary computation.
\newblock {\em PeerJ Preprints}, 6:e27315v1.

\bibitem[Goings et~al., 2012]{goings_ecology-based_2012}
Goings, S., Goldsby, H.~J., Cheng, B.~H., and Ofria, C. (2012).
\newblock An ecology-based evolutionary algorithm to evolve solutions to
  complex problems.
\newblock {\em Artificial Life}, 13:171--177.

\bibitem[Goings and Ofria, 2009]{goings_ecological_2009}
Goings, S. and Ofria, C. (2009).
\newblock Ecological approaches to diversity maintenance in evolutionary
  algorithms.
\newblock In {\em {IEEE} {Symposium} on {Artificial} {Life}, 2009. {ALife}
  '09}, pages 124--130.

\bibitem[Goldberg et~al., 1987]{goldberg_genetic_1987}
Goldberg, D.~E., Richardson, J., and Grefenstette, J.~J. (1987).
\newblock Genetic algorithms with sharing for multimodal function optimization.
\newblock In {\em Genetic algorithms and their applications: {Proceedings} of
  the {Second} {International} {Conference} on {Genetic} {Algorithms}}, pages
  41--49, Hillsdale, NJ. Lawrence Erlbaum.

\bibitem[Granger, 1969]{granger_InvestigatingCausalRelations_1969}
Granger, C. W.~J. (1969).
\newblock Investigating {Causal} {Relations} by {Econometric} {Models} and
  {Cross}-spectral {Methods}.
\newblock {\em Econometrica}, 37(3):424--438.
\newblock Publisher: [Wiley, Econometric Society].

\bibitem[Helmuth et~al., 2016]{helmuth_lexicase_2016}
Helmuth, T., McPhee, N.~F., and Spector, L. (2016).
\newblock Lexicase {Selection} for {Program} {Synthesis}: {A} {Diversity}
  {Analysis}.
\newblock In Riolo, R., Worzel, W.~P., Kotanchek, M., and Kordon, A., editors,
  {\em Genetic {Programming} {Theory} and {Practice} {XIII}}, Genetic and
  {Evolutionary} {Computation}, pages 151--167. Springer International
  Publishing.

\bibitem[Helmuth and Spector, 2015]{helmuth_general_2015}
Helmuth, T. and Spector, L. (2015).
\newblock General {Program} {Synthesis} {Benchmark} {Suite}.
\newblock In {\em Proceedings of the 2015 {Annual} {Conference} on {Genetic}
  and {Evolutionary} {Computation}}, {GECCO} '15, pages 1039--1046, New York,
  NY, USA. ACM.

\bibitem[Helmuth et~al., 2015]{helmuth_solving_2015}
Helmuth, T., Spector, L., and Matheson, J. (2015).
\newblock Solving {Uncompromising} {Problems} {With} {Lexicase} {Selection}.
\newblock {\em IEEE Transactions on Evolutionary Computation}, 19(5):630--643.

\bibitem[Hernandez et~al., 2021]{hernandez_exploration_2021}
Hernandez, J.~G., Lalejini, A., and Ofria, C. (2021).
\newblock An {Exploration} of {Exploration}: {Measuring} the ability of
  lexicase selection to find obscure pathways to optimality.
\newblock {\em arXiv:2107.09760 [cs]}.
\newblock arXiv: 2107.09760.

\bibitem[Isaac et~al., 2007]{isaac_mammals_2007}
Isaac, N. J.~B., Turvey, S.~T., Collen, B., Waterman, C., and Baillie, J. E.~M.
  (2007).
\newblock Mammals on the {EDGE}: {Conservation} {Priorities} {Based} on
  {Threat} and {Phylogeny}.
\newblock {\em PLOS ONE}, 2(3):e296.

\bibitem[Jackson, 2010]{jackson_promoting_2010}
Jackson, D. (2010).
\newblock Promoting {Phenotypic} {Diversity} in {Genetic} {Programming}.
\newblock In Schaefer, R., Cotta, C., Kołodziej, J., and Rudolph, G., editors,
  {\em Parallel {Problem} {Solving} from {Nature}, {PPSN} {XI}}, Lecture
  {Notes} in {Computer} {Science}, pages 472--481, Berlin, Heidelberg.
  Springer.

\bibitem[Kassambara, 2020]{ggpubr}
Kassambara, A. (2020).
\newblock {\em ggpubr: 'ggplot2' Based Publication Ready Plots}.
\newblock R package version 0.4.0.

\bibitem[Kauffman and Levin, 1987]{kauffman_towards_1987}
Kauffman, S. and Levin, S. (1987).
\newblock Towards a general theory of adaptive walks on rugged landscapes.
\newblock {\em Journal of Theoretical Biology}, 128(1):11--45.

\bibitem[Metevier et~al., 2019]{metevier_LexicaseSelectionGenetic_2019}
Metevier, B., Saini, A.~K., and Spector, L. (2019).
\newblock Lexicase {Selection} {Beyond} {Genetic} {Programming}.
\newblock In Banzhaf, W., Spector, L., and Sheneman, L., editors, {\em Genetic
  {Programming} {Theory} and {Practice} {XVI}}, Genetic and {Evolutionary}
  {Computation}, pages 123--136. Springer International Publishing, Cham.

\bibitem[Meyer, 2014]{infotheo}
Meyer, P.~E. (2014).
\newblock {\em infotheo: Information-Theoretic Measures}.
\newblock R package version 1.2.0.

\bibitem[Mouret and Doncieux, 2009]{mouret_overcoming_2009}
Mouret, J. and Doncieux, S. (2009).
\newblock Overcoming the bootstrap problem in evolutionary robotics using
  behavioral diversity.
\newblock In {\em Evolutionary {Computation}, 2009. {CEC}'09. {IEEE} {Congress}
  on}, pages 1161--1168. IEEE.

\bibitem[Ofria et~al., 2018]{ofria_empirical_2018}
Ofria, C., Dolson, E., Lalejini, A., Fenton, J., Jorgensen, S., Miller, R.,
  Moreno, M.~A., Stredwick, J., Zaman, L., Schossau, J., Gillespie, L., G,
  N.~C., and Vostinar, A. (2018).
\newblock Empirical.
\newblock {DOI: 10.5281/zenodo.1439475}.

\bibitem[Ofria and Wilke, 2004]{ofria_avida:_2004}
Ofria, C. and Wilke, C.~O. (2004).
\newblock Avida: {A} software platform for research in computational
  evolutionary biology.
\newblock {\em Artificial Life}, 10(2):191--229.

\bibitem[{R Core Team}, 2021]{r2021}
{R Core Team} (2021).
\newblock {\em R: A Language and Environment for Statistical Computing}.
\newblock R Foundation for Statistical Computing, Vienna, Austria.

\bibitem[Schreiber, 2000]{schreiber_MeasuringInformationTransfer_2000}
Schreiber, T. (2000).
\newblock Measuring {Information} {Transfer}.
\newblock {\em Physical Review Letters}, 85(2):461--464.
\newblock Publisher: American Physical Society.

\bibitem[Sekanina and Bidlo, 2005]{sekanina_EvolutionaryDesignArbitrarily_2005}
Sekanina, L. and Bidlo, M. (2005).
\newblock Evolutionary {Design} of {Arbitrarily} {Large} {Sorting} {Networks}
  {Using} {Development}.
\newblock {\em Genetic Programming and Evolvable Machines}, 6(3):319--347.

\bibitem[Spector, 2012]{spector_assessment_2012}
Spector, L. (2012).
\newblock Assessment of problem modality by differential performance of
  lexicase selection in genetic programming: a preliminary report.
\newblock In {\em Proceedings of the 14th annual conference companion on
  {Genetic} and evolutionary computation}, pages 401--408. ACM.

\bibitem[Tucker et~al., 2017]{tucker_guide_2017}
Tucker, C.~M., Cadotte, M.~W., Carvalho, S.~B., Davies, T.~J., Ferrier, S.,
  Fritz, S.~A., Grenyer, R., Helmus, M.~R., Jin, L.~S., Mooers, A.~O., Pavoine,
  S., Purschke, O., Redding, D.~W., Rosauer, D.~F., Winter, M., and Mazel, F.
  (2017).
\newblock A guide to phylogenetic metrics for conservation, community ecology
  and macroecology.
\newblock {\em Biological Reviews}, 92(2):698--715.

\bibitem[Tucker et~al., 2018]{tucker_relationship_2018}
Tucker, C.~M., Davies, T.~J., Cadotte, M.~W., and Pearse, W.~D. (2018).
\newblock On the relationship between phylogenetic diversity and trait
  diversity.
\newblock {\em Ecology}, 99(6):1473--1479.

\bibitem[Webb et~al., 2002]{webb_phylogenies_2002}
Webb, C.~O., Ackerly, D.~D., McPeek, M.~A., and Donoghue, M.~J. (2002).
\newblock Phylogenies and {Community} {Ecology}.
\newblock {\em Annual Review of Ecology and Systematics}, 33(1):475--505.

\bibitem[Wickham, 2016]{wickham_ggplot2:_2016}
Wickham, H. (2016).
\newblock {\em ggplot2: {Elegant} graphics for data analysis}.
\newblock Springer-Verlag New York.

\bibitem[Yao and Li, 2020]{yao_EffectiveTransferEntropy_2020}
Yao, C.-Z. and Li, H.-Y. (2020).
\newblock Effective {Transfer} {Entropy} {Approach} to {Information} {Flow}
  {Among} {EPU}, {Investor} {Sentiment} and {Stock} {Market}.
\newblock {\em Frontiers in Physics}, 8:206.

\end{thebibliography}

\end{document}